\documentclass[sigconf,authorversion]{acmart}
\AtBeginDocument{%
  }
\usepackage{ragged2e}
\usepackage{microtype}
\usepackage[caption=false]{subfig}
\usepackage{balance}
\justifying

\usepackage{subcaption}
\usepackage{ragged2e}

\acmDOI{10.1145/3776734.3794350}
\acmYear{2026}
\copyrightyear{2026}
\acmISBN{979-8-4007-2321-6/2026/03}
\acmConference[HRI Companion '26]{Companion Proceedings of the 21st ACM/IEEE International Conference on Human-Robot Interaction}{March 16--19, 2026}{Edinburgh, Scotland, UK}
\acmBooktitle{Companion Proceedings of the 21st ACM/IEEE International Conference on Human-Robot Interaction (HRI Companion '26), March 16--19, 2026, Edinburgh, Scotland, UK}
\received{2025-12-08}
\received[accepted]{2026-01-12}

\begin{document}

\title{Co-designing a Social Robot for Newcomer Children’s Cultural and Language Learning
}

\author{Neil Fernandes}
\authornote{Emails: \texttt{\{neil.fernandes, yue.hu, kerstin.dautenhahn\}@uwaterloo.ca}}

\orcid{0009-0003-5043-9353}
\affiliation{%
  \institution{University of Waterloo}
  \city{Waterloo}
  \state{Ontario}
  \country{Canada}
}

\author{Tehniyat Shahbaz}
\authornote{Emails: \texttt{tehniyat.shahbaz@gmail.com, edrobinson@unitedforliteracy.ca}}
\orcid{0009-0002-5736-7105}
\affiliation{%
  \institution{United for Literacy}
    \city{Toronto}
  \state{Ontario}
  \country{Canada}
  }



\author{Emily Davies-Robinson}
\authornotemark[2]
\orcid{0009-0006-9762-8997}

\affiliation{%
 \institution{United for Literacy}
  \city{Toronto}
  \state{Ontario}
  \country{Canada}}

\author{Yue Hu}
\authornotemark[1]

\orcid{0000-0002-3846-9096}
\affiliation{%
  \institution{University of Waterloo}
  \city{Waterloo}
  \state{Ontario}
  \country{Canada}}

\author{Kerstin Dautenhahn}
\authornotemark[1]
\orcid{0000-0002-9263-3897}
\affiliation{%
  \institution{University of Waterloo}
  \city{Waterloo}
  \state{Ontario}
  \country{Canada}}

\renewcommand{\shortauthors}{Fernandes et al.}

\begin{abstract}
Newcomer children face barriers in acquiring the host country’s language and literacy programs are often constrained by limited staffing, mixed-proficiency cohorts, and short contact time. While Socially Assistive Robots (SARs) show promise in education, their use in these socio-emotionally sensitive settings remains underexplored. This research presents a co-design study with program tutors and coordinators, to explore the design space for a social robot, \textit{Maple}. We contribute (1) a domain summary outlining four recurring challenges, (2) a discussion on cultural orientation and community belonging with robots, (3) an expert-grounded discussion of the perceived role of an SAR in cultural and language learning, and (4) preliminary design guidelines for integrating an SAR into a classroom. These expert-grounded insights lay the foundation for iterative design and evaluation with newcomer children and their families.
\end{abstract}




\maketitle

\section{Introduction}

Newcomer children often need to acquire the language of their host country while simultaneously adjusting to unfamiliar school systems, cultural norms, and social expectations \cite{yang2024towards}. In this paper, we use the term `newcomer children' to refer to young children from immigrant or refugee backgrounds who have migrated to Canada within the past five years and are in the process of learning English as an additional language, adapting prior definitions of newcomer youth and students used in Canadian education research \cite{kalchos2022access,IRCC2018}. Community based literacy programs provide after school support through worksheets and small group activities. Tutors in these settings must support culturally and linguistically diverse groups where children differ in age, first language, country of origin, and proficiency. As a result, they must balance language and socio-emotional support as well as classroom management, often without information about each learner's proficiency. SARs have been proposed as one way to support learning in such constrained environments. Within Robot-Assisted Language Learning (RALL), social robots have shown to increase motivation and confidence \cite{kanda2004interactive}, support speaking practice \cite{kory2019assessing}, and scaffold early literacy skills, outperforming virtual agents or tablets \cite{konijn2022social}. Social robots are often perceived as more affiliative and less judgmental than screen based systems \cite{caruana2023perceptions}, which may be particularly valuable for learners who are apprehensive about making mistakes in front of adults or peers \cite{li2015benefit,holthower2023robots}. Work that explicitly considers newcomer populations is beginning to emerge \cite{lugrin2018social,louie2022designing,yang2024towards}, but existing studies are typically short term, rarely situated in community programs.

This paper addresses this gap using a co-design approach with tutors and coordinators from United for Literacy (UFL) - a national charitable literacy organization that delivers programs to children, youth and adults in urban, rural and remote communities across Canada - who have direct insight into learners' needs and the constraints they face as tutors \cite{steen2013co}. Using the Design Thinking process \cite{brown2008design} and a co-design approach \cite{iniesto2022review,greenhalgh2016achieving,alves2022robots,sanders2008co}, we conducted a co-design meeting wherein experts reflected on their everyday practice and reacted to a robotic prototype that we call \textit{Maple}. This paper presents four primary contributions: (1) a domain summary that describes four recurring challenges in supporting newcomer children, (2) a discussion of how cultural orientation and community belonging emerge in experts' accounts, (3) an analysis of the perceived roles of an SAR in this context, and (4) preliminary design guidelines for a knowledgeable peer-like robot that addresses these challenges while aligning with current pedagogies.

\section{Methods and Procedures}

\subsection{Co-designers}
Our design team comprised four participants: two expert second-language (L2) tutors\footnote{co-authors in paper} from UFL that delivers literacy programs\footnote{\url{https://www.unitedforliteracy.ca/Programs}}
and two Human–Robot Interaction (HRI) researchers (one professor and one graduate student). One L2 expert was a regional manager overseeing the organization's programs in the Region of Waterloo with seven years of experience, and the other L2 expert was a Program Support Staff in the region who has been with UFL for two and a half years as both staff and volunteer.

\subsection{Robotic System Overview}

\begin{figure}[h]
    \centering
    \includegraphics[width=0.75\linewidth, alt={Color photograph of the Maple robot prototype standing on a table. A smartphone mounted as the head shows a cartoon face with large eyes and a small smile, while the compact black robot body has articulated arms, one raised as if waving. The image highlights the robot’s small size and friendly, approachable appearance.}]{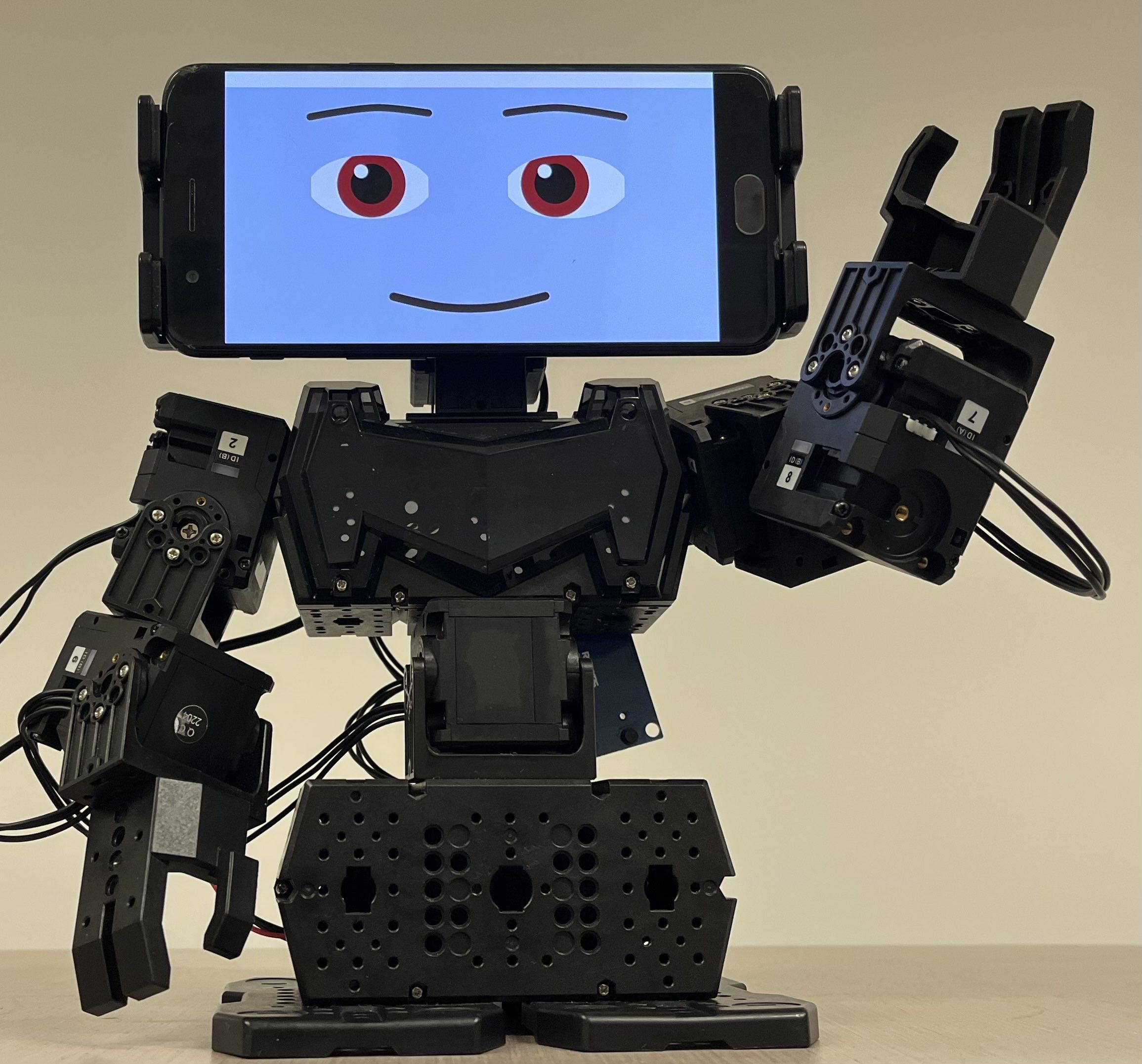}
    \caption{The \textit{Maple} robot prototype.}
    \label{maple}
\end{figure}

Our prototype, \textit{Maple} (see \autoref{maple}), is assembled from the ROBOTIS Engineer Kit~1\footnote{\url{https://en.robotis.com/model/page.php?co_id=prd_engineerkit}}, previously evaluated in HRI work \cite{yang2024towards}. The platform uses multiple DYNAMIXEL servos to obtain expressive upper-body gestures and head movements, and is configured as a compact table-top humanoid (to support coordinated verbal/non-verbal interaction). The robot is adapted to mount a smartphone as the face display, allowing non-verbal gestures to be combined with animated eyes and a moving mouth for social cues.

The control stack is implemented in Python on the Robotics Operating System (ROS) Noetic Ninjemys distribution, using a modular architecture with separate components for motion control, audio output, facial animation, and higher-level behavior tree  \cite{colledanchise2018behavior}. Facial animation is provided by PyLips \cite{dennler2024pylips}, which synchronizes mouth movements and expressions with robot speech, and the overall scheduling of speech and motion follows principles from RoboSync \cite{tang2023robosync}. To make interaction flows easy to create and modify, we provide a web-based user interface to be used on a tablet, built in React, similar to \cite{ivanov2021online} that connects to the robot via a ROS bridge \cite{crick2012ros}, allowing tutors and researchers to define story-based activities and quizzes without changing the underlying control code.

\subsection{Procedure and Analysis}
We conducted one 60-minute online co-design meeting with the participants as a group over Microsoft Teams. The meeting followed a semi-structured, interview-style format and was recorded and transcribed. In the first part of the session, experts reflected on their everyday practice and current challenges in the community language learning program. In the second part, the experts watched a demo of a suggested interaction with the working \textit{Maple} prototype and brainstormed possible activities based on the robot's capabilities. The data was then summarized using the domain summary method \cite{braun2006using}, which resulted in the themes that capture recurring challenges and pedagogical goals, which formed the basis for the preliminary guidelines.

\section{Preliminary Results}

\subsection{Challenges}
Across the first co-design meeting, experts described four recurring challenges: (1) Language Barriers and Skill Levels, (2) Lack of Motivation and Attention., (3) Absence of Formal Proficiency Baselines, and (4) Lack of One-on-One Attention and Support.

\subsubsection{Language Barriers and Skill Levels}
Language barriers were consistently identified as an immediate difficulty. The recruitment of learners is very inclusive, and before sessions start tutors will not have prior knowledge of the children's individual language proficiencies. Learners are highly diverse, originating from many language backgrounds with variable English proficiency. In some cases, children ``\textit{[...] do not speak a single word of English [...]},'' meaning tutors cannot rely on a shared language for rapport or task explanation. Tutors, often not speaking the learners' first languages, rely on translation tools to go back and forth with the child. In this case, Tutors informally gauge literacy levels by reading with children and pausing at difficult words. These efforts are further complicated by inconsistent attendance across the term. Experts described planning new strategies for particular learners, only to find that \textit{``[...] some children attend as few as three out of eight sessions [...]}. This irregular attendance makes it difficult to implement new approaches consistently. Attendance \textit{``can be impacted by a multitude of factors, often which are out of our control, such as transportation issues, parental availability for drop-off and pick-up [...]''}. As a result, tutors must continually adapt activities to a group whose composition and skill levels can vary from week to week.

\subsubsection{Lack of Motivation and Attention}
Sustaining learner attention and motivation was a second recurring challenge. Program activities rely heavily on paper-based worksheets, which tutors observed are often ``\textit{hard for learners to just learn through worksheets}'' alone, especially regarding abstract grammar. Learners may disengage, or lose interest with the material's relevance. Recognizing that learners have already spent a full day in school before attending the program, one expert explained, ``\textit{[...] we like to include games so it feels less like school. Of course, we want to ensure that sessions are educational, but also making sure it's fun!''} To mitigate disengagement, tutors therefore interject playful, interactive games, described by one expert as ``\textit{helpful for re-engaging learners''}. This was especially emphasized for the younger learners, ``\textit{especially for the younger learners, they rather do activities that involve standing up and moving''}. The experts also noted that attention difficulties are not solely task-related, reporting perceived ``\textit{[...] shorter attention spans than they are used to [...]}'' and challenges arising from, for example, Attention Deficit Hyperactivity Disorder (ADHD) or Autism Spectrum Disorder (ASD). Even if children have been formally diagnosed, tutors will not be aware of the diagnoses.


\subsubsection{Absence of Formal Proficiency Baselines}
The third challenge is the lack of clear, consistent proficiency baselines. The program's open intake includes no formal L2 assessment, so tutors often \textit{``[...] do not know the english proficiency levels [...]''} of newcomer children. Tutors rely on informal strategies, like reading a text with the child, to ascertain a starting point. Learners span the full range from pre-acquisition to having several years of English study. This semi-formal-assessment is fragile due to variable attendance from both children and tutors. This lack of a shared baseline complicates activity differentiation and progress tracking.

\subsubsection{Lack of One-on-One Attention and Support}
Finally, the structural difficulty of providing one-on-one attention in a group setting was discussed. Typical sessions have low tutor-to-learner ratios, where tutors must \textit{``[...] float around [...]''} the room and shift their attention, rather than dedicating time to a single child, which makes responding to individual needs difficult. Experts suggest that this is especially visible for children with low English proficiency, ``\textit{[...] who require one-on-one support [...]}''. Sometimes the literacy program takes in same day sign-ups for newcomer children, which further reduces opportunities for rapport-building. Despite staff's best efforts, systemic constraints make sustained individual attention to the learners challenging. These four challenges formed the backdrop for conceptualizing the robot's role.

\subsection{Cultural Orientation and Community Belonging}
Beyond language, experts repeatedly framed the language program as a space for cultural orientation and community building. One expert described families as \textit{`` [...] trying to get acclimatized or trying to get familiarized with the Canadian Culture.''}. In many cases,\textit{ ``[...] parents don’t even know [these cultural] norms, especially if they’ve only arrived to Canada very recently [...]''}, so the sessions help the family navigate norms of their host country.

\begin{figure}[!h]
    \centering
    \includegraphics[width=1\linewidth, alt={Six panel storyboard with Canadian animal characters illustrating a short vocabulary lesson about politeness on a city bus.
}]{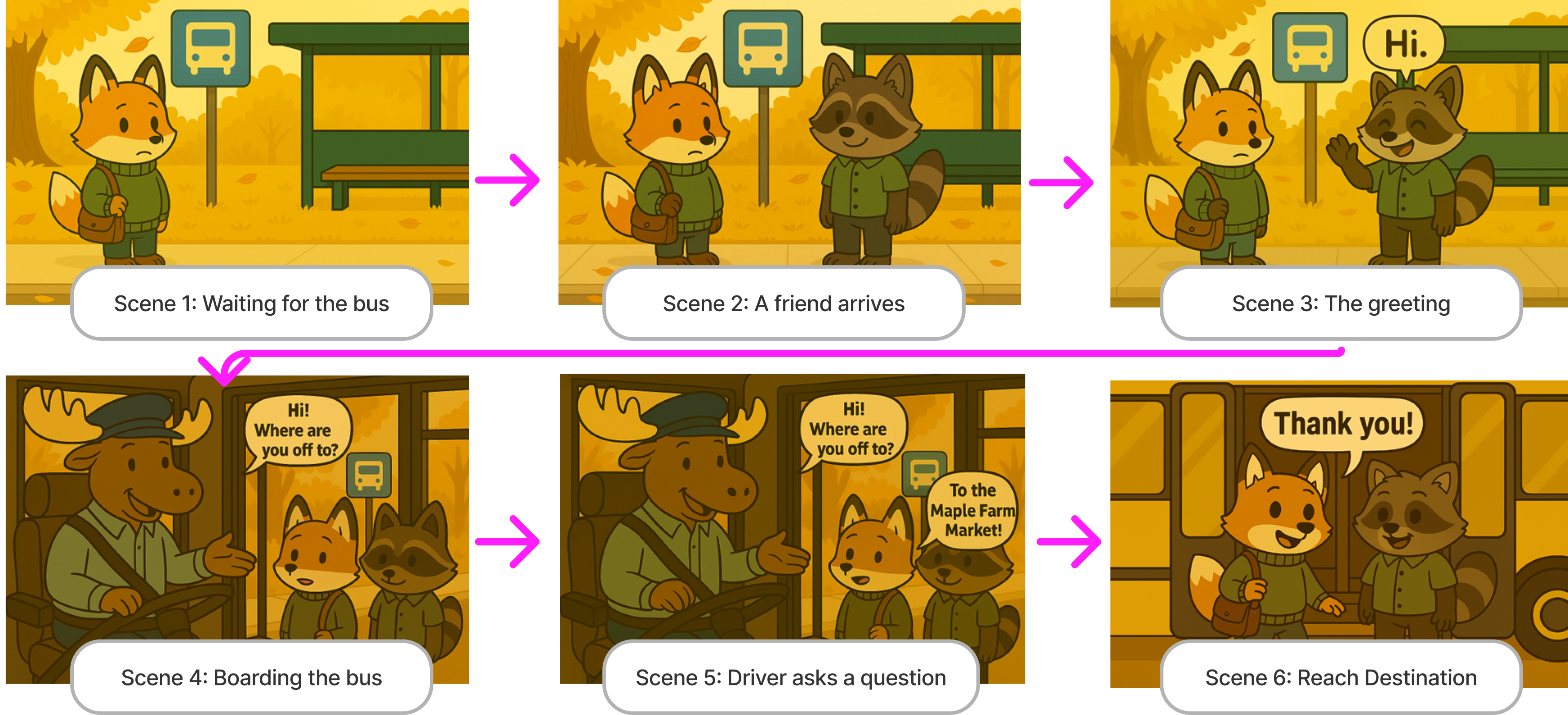}
    \caption{Suggested storyboard with vocabulary learning, grounded in a storyboard with animals. Key point of being polite to people around you (in this case, the bus driver)}
        \label{fig:storyboard}
\end{figure}

Tutors also highlighted activities that explicitly embed cultural content from Canada. For example, one popular activity is \textit{``[...]a pop-up shop. So we have play money and then we create [...] a market and they have to purchase [...] buttons and [...] pencils,''} and \textit{``[...] we assign monetary values to it and this way they learn about Canadian money [...] like what is a Loonie? what’s a Toonie? Like why is it called a Toonie\footnote{A Loonie and Toonie are informal ways of referring to One and Two Canadian Dollar coins, see https://educacentre.com/loonie-and-toonie/}}?'' Similarly, when discussing the story world, experts suggested using \textit{``Animals like Raccoons and Moose''}, to ground vocabulary in recognizable references. Cultural norms around interaction were also discussed (an example of this can be seen in \autoref{fig:storyboard}, images generated by GPT 4o \cite{HelloGPT4o}). In one example, experts discussed a school scenario where \textit{``[...] greeting friends or other children of a similar age with hi is probably, you know, I mean that's, what children do [...],''} but \textit{``[...] if they go to school and they greet the teacher with hi, then, they should probably say good morning, or you know, good afternoon [out of respect].''} These comments position \textit{Maple} not only as a language-learning peer, but also as a potential cultural bridge: a robot that can help children practice everyday cultural concepts and norms.

\subsection{Perceived Social Role of \textit{Maple}}
\begin{figure}[h!]
    \centering
    \includegraphics[width=0.91\linewidth, alt={Drawing illustration of a tutor, a child, and the robot seated together at a table in a classroom. The tutor sits on the left, leaning toward the child with an encouraging expression, while the child works on a worksheet and looks toward the robot. The robot is on the right side of the table facing the child, showing a triadic interaction where the robot acts as a supportive peer alongside the human tutor.}]{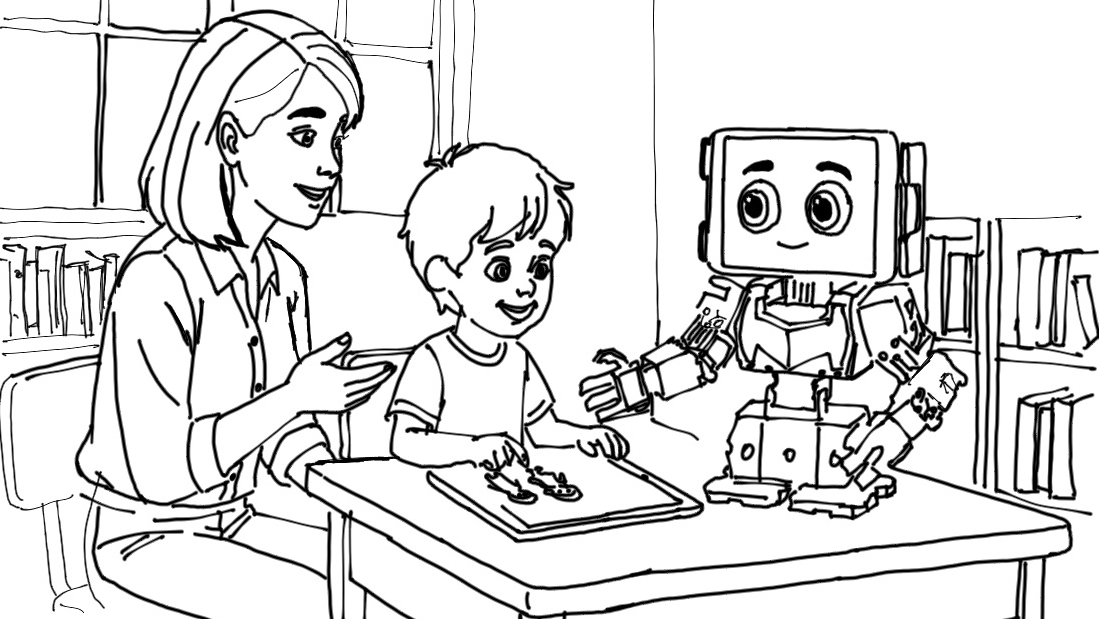}
    \caption{Illustration showing a potential triadic ‘tutor–child–robot’
interaction}
    \label{triadeteaser}
\end{figure}

Across the discussion, experts converged on \textit{Maple} as a peer-like companion rather than a teacher or authority figure. In contrast adult tutors who may sometimes appear  intimidating, \textit{Maple} was described as a `softer', more approachable presence to encourage participation. Experts stressed that \textit{Maple} should not replace teachers, but act as a peer within a triadic `tutor–child–robot' interaction: seated beside the child, taking on conversational roles, while the tutor monitors, intervenes as needed, and moves between learners. This is illustrated in \autoref{triadeteaser}. A peer role, they noted, requires matching behaviors—friendly and collaborative task-sharing, with deference to the tutor on complex topics. Even when supporting assessment, \textit{Maple} should remain a supportive peer rather than a teacher.

\subsection{Preliminary Design Guidelines}
Derived from the challenges and experts’ discussions, we outline a set of preliminary design guidelines for a peer SAR robot that fits the literacy program's needs.

\subsubsection{Guideline 1: Provide multi-modal and multilingual scaffolding for language barriers.}
The robotic system should present key vocabulary and instructions in multiple modalities, for example robot speech, on-screen text, and simple gestures.
When possible, important words in digital stories  should appear in both English and the learner’s first language, with the robotic system repeating the English term and visually pointing or gesturing toward the corresponding image.
This could directly address the initial language barrier and help children follow along even when their English is only emerging.

\subsubsection{Guideline 2: Use short, story-based activities to support attention and motivation.}
Drawing on Social Learning Theory \cite{bandura1977social} and the Audiolingual method \cite{zillo1973birth, zhou2015educational, valdman1970toward}, the system could address fluctuating attention by organizing interactions into story-driven segments where the robot systematically repeats and reinforces target words with the young child from a story-based activity on a tablet. Turn-taking narration, where the robotic system and the tutor alternate lines or paragraphs, could make the activity feel more playful while keeping the learning goal in focus. The robotic system’s existing gestures, such as waving and `I am happy' acknowledgments, could be used to celebrate small achievements and redirect learners to the activity when their focus drifts away from the lesson content.

\subsubsection{Guideline 3: Learn a second language \textit{through} cultural orientation}
In the co-design meeting, experts suggested using \textit{Maple} to rehearse everyday situations in the host community. Drawing on work in language socialization \cite{schieffelin1986language}, we interpret such activities as opportunities for newcomers to be socialized \emph{through} language and \emph{to} use language appropriately in their communities, by participating in meaningful routines rather than isolated drills \cite{duranti_ochs_schieffelin_2012_handbook,garrett_baquedano_lopez_2002}. Based on these discussions, \textit{Maple}’s activities could be implemented as short role-play scenes in a story that mirror common routines (e.g. meeting children at school to be friends with). Within these scenes, the robot should model key social interactions (e.g. greeting or politely declining), while tutors remain available to step in or adapt the scenario. 

\subsubsection{Guideline 4: Embed formal assessments into playful interactions.}
Due to the absence of consistent proficiency baselines, the robotic system could help collect simple indicators of reading and language level without the activity turning into a formal test.
For example, the robot could ask the child to read a short sentence from a story or to choose a correct word on the screen, and then log whether the child completed the action independently or with support.
These micro-assessments can be built into story-like activities and could produce data that tutors can easily track to adjust future worksheets and groupings.

\subsubsection{Guideline 5: Support one-on-one attention within a triadic setup.}
To address limited one-on-one attention, \textit{Maple} can be designed for triadic interaction in which a tutor, a learner, and the robot jointly share attention. In this model, a tutor circulates between several children while \textit{Maple} stays with one learner, guiding them through a short reading task. The interaction flow can include points where the tutor steps in, (e.g.\ by asking a brief quiz). This keeps the tutor involved and preserves what one expert called the \textit{`` [...] human-ness aspect of the tutor's interaction [...]''} within the overall experience including \textit{Maple}. Design choices should also consider fairness, for instance by rotating learners to interact with \textit{Maple} or sometimes placing the robot at the center of a small group activity \textit{``[...] so that no child feels overlooked or excluded.''}. Each learner could also be assigned their `own' \textit{Maple} robot.

\section{Discussion and Future Work}
The four challenges identified above point to complementary tasks a robot could perform alongside human tutors, which we summarize in five preliminary design guidelines. Tutors’ descriptions of using online translation tools, improvising games, and conducting informal reading checks suggest micro-roles for the robot such as offering multilingual and visual explanations, repeating target phrases, stepping through short reading tasks that implicitly serve as formal assessments, and gently redirecting attention during story segments. We therefore frame the robot’s primary value as both a \textbf{tutor-support tool} and a \textbf{child-peer tool} that helps manage constraints in the program: acting as an \textit{``attention load-balancer''} by handling repetitive scaffolding tasks \cite{bliss1996effective} and supporting individual children while tutors rotate between learners (Guidelines 1 and 5), sustaining engagement through short story-based activities (Guideline 2), embedding everyday cultural routines and local practices into those stories so that children practice English \textit{through} cultural orientation (Guideline 3), and gathering low-level proficiency data during these interactions to help tutors establish baselines without adding formal tests (Guideline 4).


\section{Conclusion}

This late-breaking report presents an expert co-design study with UFL staff to inform Maple, a peer-like socially assistive robot for newcomer children’s language learning and cultural orientation in community programs. We identify recurring challenges and derive preliminary design guidelines that position the robot within a triadic tutor-child-robot interaction to complement human instruction. As these guidelines are expert-grounded, ongoing work will validate and refine them through in-situ ``read with me'' story sessions in UFL workshops, alongside tutor and parent feedback.

\begin{acks}
    Thank you to Cheng Tang, Nathan Dennler and Ali Yamini for their help on \textit{Maple}. This research was undertaken, in part, thanks to funding from the Canada 150 Research Chair Program.
\end{acks}


\bibliographystyle{ACM-Reference-Format}
\bibliography{citations}

\end{document}